%% file: main.tex
\documentclass{article}

\usepackage{iclr2027_conference,times}
\input{math_commands.tex}

\usepackage{hyperref}       
\usepackage{url}            
\usepackage{multicol}
\usepackage{multirow}
\usepackage{graphicx}

\usepackage[utf8]{inputenc} 
\usepackage[T1]{fontenc}    
\usepackage{booktabs}       
\usepackage{amsfonts}       
\usepackage{nicefrac}       
\usepackage{microtype}      
\usepackage{xcolor}         

\usepackage{amsmath, amssymb}
\usepackage[capitalise]{cleveref}
\usepackage{bm}
\usepackage{colortbl}
\usepackage{epsfig}
\usepackage{multirow}
\usepackage{tabularx}
\usepackage{enumitem}
\usepackage{tabularx}
\usepackage{booktabs}
\usepackage{array}
\usepackage[table]{xcolor}
\newcommand{\name}{AtomicRL}
\definecolor{mygray}{gray}{.95}
\definecolor{myblue}{HTML}{ECF2F8}
\definecolor{mycyan}{cmyk}{.1,0,0,0}
\definecolor{mygray}{gray}{.95}
\definecolor{mypink}{rgb}{.99,.91,.95}
\definecolor{mypurple}{rgb}{0,0,0}
\newcolumntype{C}{>{\centering\arraybackslash}X}
\usepackage[dvipsnames]{xcolor}
\usepackage{subcaption}
\newcommand{\mypara}[1]{\vspace{1mm}\noindent\textbf{#1}}
\newsavebox{\calvinTableBox}
\newsavebox{\liberoNaturalBox}

\usepackage{booktabs}
\usepackage{threeparttable}
\usepackage{array}

\usepackage{textcomp, gensymb}
\usepackage{caption}
\usepackage{makecell}
\usepackage{wrapfig}
\usepackage{pifont}

\definecolor{mycyan}{cmyk}{.1,0,0,0}
\definecolor{mygray}{gray}{.95}
\definecolor{mypink}{rgb}{.99,.91,.95}

\title{\name{}: Atomic Reinforcement Learning for Long-Horizon Robotic Manipulation}

\author{
    \textbf{Likui Zhang$^{1}$, Zhe Zhu$^{1}$, Zhihao Zhan$^{1}$, Yibo Bai$^{3}$, Hanlin Wang$^{2}$, Dongwei Li$^{3}$, } \\[3pt]
    \textbf{ Bin Wang$^{3}$, Tianle Zhang$^{3}$,
    Xiaodan Liang$^{1\dagger}$, Liang Lin$^{1\dagger}$} \\[4pt]
    $^1$Sun Yat-sen University 
    $^2$The Hong Kong Polytechnic University  $^3$JD Explore Academy 
}

\iclrfinalcopy 
\begin{document}

\maketitle
\fancyhead{}
\begingroup
\renewcommand{\thefootnote}{\fnsymbol{footnote}}
\footnotetext[2]{Co-corresponding authors.}
\endgroup

\input{sec/0_abstract}    
\input{sec/1_intro}

\input{sec/2_related_work}
\input{sec/3_methodology}
\input{sec/4_experiment}

\input{sec/5_conclusion}

{
    \small
    \bibliographystyle{iclr2027_conference}
    \bibliography{main}
}

\input{sec/X_suppl}

\end{document}

%% file: math_commands.tex
\usepackage{amsmath,amsfonts,bm}

\def\eqref#1{equation~\ref{#1}}

\def\1{\bm{1}}

\DeclareMathAlphabet{\mathsfit}{\encodingdefault}{\sfdefault}{m}{sl}
\SetMathAlphabet{\mathsfit}{bold}{\encodingdefault}{\sfdefault}{bx}{n}



%% file: sec/0_abstract.tex
\begin{abstract}

Vision–language–action (VLA) models have shown promise in robotic manipulation, but imitation learning (IL) alone often struggles to master complex long-horizon skills. Reinforcement learning (RL) offers an effective avenue for further improving VLA policies. However, existing VLA-RL methods typically suffer from delayed and biased credit assignment due to trajectory-level sparse rewards, while imbalanced subtask mastery creates bottlenecks that limit overall performance. To address these challenges, we propose AtomicRL, an RL post-training framework that leverages atomic task structures for reliable credit assignment and adaptive exploration. AtomicRL introduces Semantic-Aware Policy Optimization (SAPO), which uses semantic-aware gating to constrain TD residuals from failed atomic subtasks from propagating across task boundaries, mitigating credit contamination while preserving intra-subtask credit assignment and value bootstrapping. In addition, Bottleneck-Adaptive Target Exploration (BATE) dynamically coordinates task sampling and local exploration incentives based on the mastery of individual atomic subtasks, directing learning toward the remaining bottlenecks. Finally, we develop a Real2Sim2Real pipeline that bridges online RL in simulation with real-world deployment, facilitating practical transfer to physical robots. On LIBERO-LONG, AtomicRL achieves absolute performance gains of 3.6\% and 4.0\% over PPO across two VLA architectures. On CALVIN ABC-D, it improves full-task performance by up to 9.5\% over PPO. Combined with the Real2Sim2Real pipeline, AtomicRL further improves real-world task performance by an average of 21.3\%. These results demonstrate that AtomicRL effectively improves the learning efficiency and manipulation performance of VLA policies on long-horizon tasks.

\end{abstract}

%% file: sec/1_intro.tex
\section{Introduction}

Recent advances in Vision-Language Models (VLMs)~\citep{beyer2024paligemmaversatile3bvlm, deitke2024molmopixmoopenweights, yang2025qwen3technicalreport} have established Vision–Language–Action (VLA)~\citep{intelligence2025pi05visionlanguageactionmodelopenworld,kim2025finetuningvisionlanguageactionmodelsoptimizing, liu2025rdt1bdiffusionfoundationmodel} models as a prominent paradigm for robotic manipulation. Although imitation learning (IL) from large-scale robot demonstrations enables VLA models to acquire diverse manipulation skills, its performance is inherently constrained by the coverage of the demonstration data. Reinforcement learning (RL) overcomes this limitation by directly optimizing policies with task feedback obtained through environment interaction, and recent work~\citep{li2026simplevla,xu2026rltokenbootstrappingonline} has successfully applied RL to VLA models. However, existing VLA-RL methods still face substantial challenges in long-horizon manipulation tasks.

First, most methods rely on terminal-only success rewards. In an episode that completes the first $k$ subtasks but fails at subtask $k+1$, the absence of a terminal success reward causes generalized advantage estimation (GAE) to propagate failure-related temporal-difference (TD) residuals backward to $t_0$, suppressing the advantages of correctly executed actions. We refer to this as credit contamination, where useful behavior in failed trajectories is systematically devalued, with the effect worsening as the horizon grows.
Prior work~\citep{chen2025robohorizonllmassistedmultiviewworld, yang2026actorcriticcontinuousactionchunks, tan2025robodopaminegeneralprocessreward, zhai2025visionlanguageactioncriticmodelroboticrealworld} mitigates reward sparsity through stage-level preference optimization or subtask-level intermediate rewards, but these approaches alter reward density without resolving credit contamination. 
Second, subtask mastery is uneven, with failures increasingly concentrated in a few bottleneck subtasks during later training. Uniform task sampling and fixed exploration then waste interaction budget on saturated subtasks while providing insufficient exploration for the bottlenecks.

\label{sec:intro}
\begin{figure}[t]
    \centering
    \vspace{-10pt}
    \includegraphics[width=0.9\textwidth]{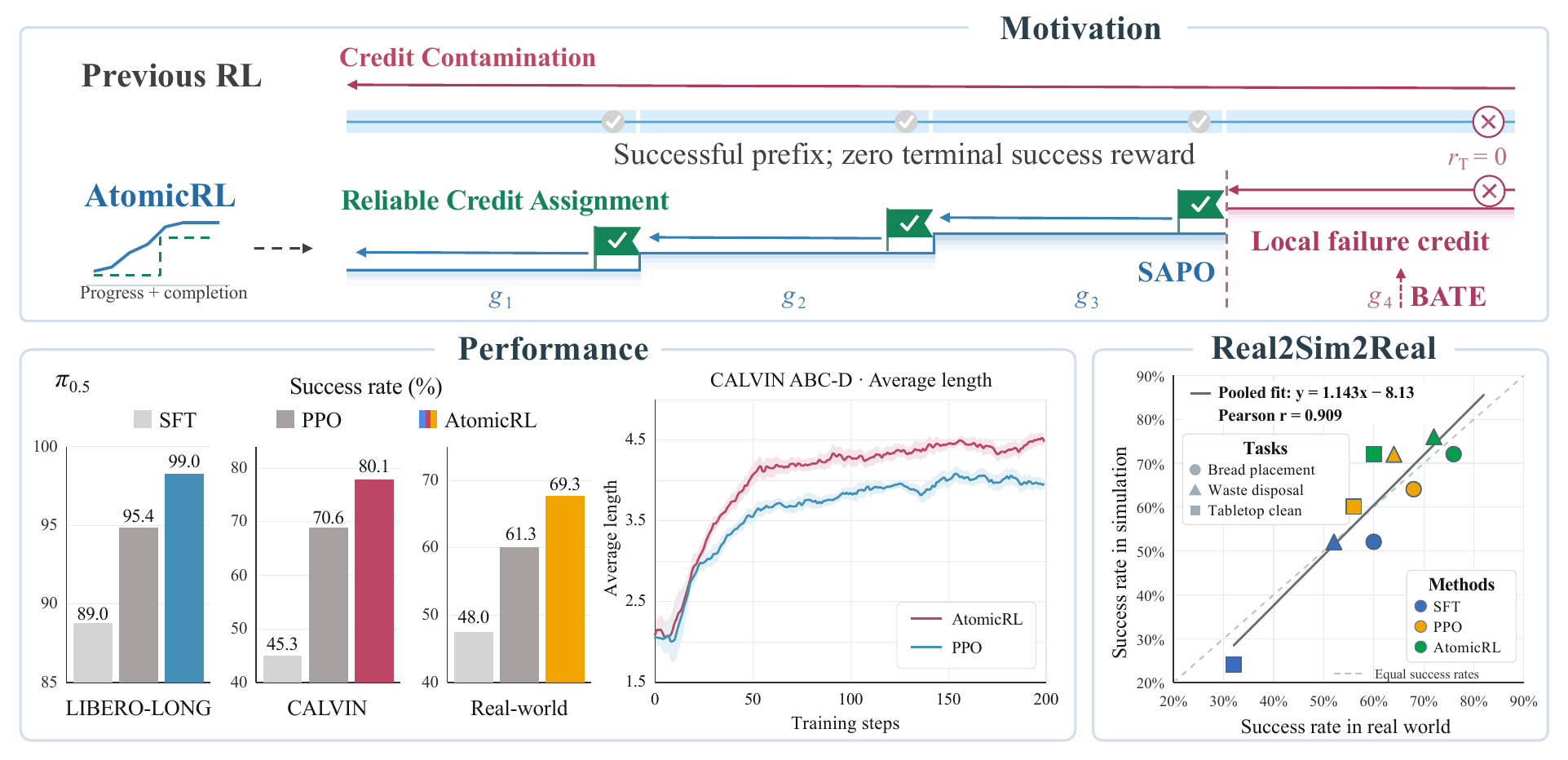} 
    \vspace{-6pt}
    \caption{\textbf{Overview of AtomicRL.} Unlike previous RL methods that use standard GAE which may suffer from credit contamination, AtomicRL uses SAPO to promote more reliable credit assignment. Together with BATE and the Real2Sim2Real pipeline, AtomicRL achieves strong performance on long-horizon manipulation tasks in both simulation and the real world.}
    \label{teaser} 
    \vspace{-15pt} 
\end{figure}

To address these challenges, we propose AtomicRL, as illustrated in Fig.~\ref{teaser}, an RL post-training framework built around atomic task structure. Our key insight is that semantic subtasks provide a unified basis for both the propagation of learning signals and the allocation of exploration resources. AtomicRL couples Semantic-Aware Policy Optimization (SAPO) with Bottleneck-Adaptive Target Exploration (BATE), supported by a lightweight Progress–Completion Critic (PCC) that estimates atomic subtask progress and completion boundaries. PCC also provides progress estimates for constructing dense progress-based shaping rewards (PBSR). 
SAPO modulates TD-residual propagation in advantage estimation with a soft gate conditioned on semantic boundaries. Within each atomic subtask, it preserves standard GAE trace decay; across subtask boundaries, it maintains propagation between successful subtasks while attenuating residuals from failed subtasks. By leaving one-step TD residuals and value bootstrapping unchanged, SAPO reduces failure interference while preserving intra-subtask credit assignment and successful subtask dependencies.
BATE tracks the mastery of each subtask via a moving average of its completion rate, jointly adjusts the sampling weight of full tasks and the strength of subtask-level entropy regularization, directing interaction and exploration resources toward bottleneck subtasks.
Finally, to support physical deployment, we develop an automated Real2Sim2Real pipeline. Given a single real-world multimodal demonstration and the required auxiliary information, the pipeline reconstructs an interactive simulation environment. It connects simulation-based RL with real-robot deployment, shifting costly physical trial-and-error to parallel simulation with reproducible resets.

We conduct extensive experiments in both simulation and the real world. On LIBERO-LONG~\citep{liu2023liberobenchmarkingknowledgetransfer} AtomicRL achieves success rates of 99.0\% with $\pi_{0.5}$~\citep{intelligence2025pi05visionlanguageactionmodelopenworld} and 98.8\% with OpenVLA-OFT~\citep{kim2025finetuningvisionlanguageactionmodelsoptimizing}, improving over PPO by 3.6\% and 4.0\%. On CALVIN ABC-D~\citep{mees2022calvinbenchmarklanguageconditionedpolicy}, the average completed sequence length improves from 3.56 to 4.56, while the success rate of completing full five-subtask chain improves by 9.5\% over PPO.
Combined with the Real2Sim2Real pipeline, AtomicRL further yields an average improvement of 21.3\% on real-robot long-horizon tasks. These results demonstrate the effectiveness of AtomicRL for long-horizon VLA reinforcement learning.  Overall, our contributions are as follows:
\begin{itemize}
    \item We introduce AtomicRL, an RL post-training framework built around atomic task structure for long-horizon robotic manipulation.

    \item We propose SAPO to mitigate cross-subtask credit interference and BATE to adaptively direct task sampling and exploration toward bottleneck subtasks.

    \item We develop a Real2Sim2Real pipeline that streamlines simulation construction and RL training for efficient real-robot deployment.
    \item We validate the effectiveness of AtomicRL through extensive experiments conducted in both simulated environments and real-world robots.
\end{itemize}


%% file: sec/2_related_work.tex
\section{Related Work}
\label{sec:related_work}

\subsection{Reinforcement Learning  for VLA Models}

Reinforcement learning (RL) optimizes policies through environment interaction and task-level feedback. Recent work has explored both offline and online RL paradigms for VLA policies. Offline RL approaches leverage task feedback to enable continued policy improvement~\citep{huang2025corftefficientfinetuningvisionlanguageaction, zhang2025grapegeneralizingrobotpolicy, zhang2025reinbotamplifyingrobotvisuallanguage}. $\pi_{0.6}^{*}$~\citep{intelligence2025pi06vlalearnsexperience} integrates expert demonstrations, autonomous interaction data, and human corrections through advantage-conditioned learning. For scalable online training, VLA-RL~\citep{lu2025vlarlmasterfulgeneralrobotic} and SimpleVLA-RL~\citep{li2026simplevla} explore frameworks for continually improving VLA policies through environment interaction, while RL4VLA~\citep{liu2026rlbringvlageneralization} systematically investigates how different RL algorithms affect VLA generalization. Furthermore, Flow Q-Learning~\citep{park2025flowqlearning}, Flow-GRPO~\citep{liu2025flowgrpotrainingflowmatching} and $\pi_{RL}$~\citep{chen2026pitextttrlonlinerlfinetuning} investigate online RL fine-tuning for flow-matching policies, addressing challenges in action likelihood computation and exploration.

RL for long-horizon manipulation introduces additional challenges due to sparse and delayed task feedback~\citep{chen2026sarmstageawarerewardmodeling, li2025grrlgoingdexterousprecise, dalal2024planseqlearnlanguagemodelguided}. STARE-VLA~\citep{xu2025starevlaprogressivestageawarereinforcement} exploits the semantic stage structure of trajectories to perform stage-level preference optimization and reward shaping, while IG-RFT~\citep{su2026igrftinteractionguidedrlframework} combines trajectory-level and subtask-level rewards to provide intermediate guidance. However, dense reward shaping provides useful process guidance over long-horizon trajectories, but cannot eliminate the spurious interference caused by later-stage failures.
In contrast, AtomicRL gates the propagation of advantage signals at the semantic boundaries of atomic tasks to prevent cross-task interference from unreliable signals, while allocating the exploration budget to bottleneck atomic tasks to improve the efficiency of long-horizon RL.

\subsection{Online Reinforcement Learning in Real World}

Online reinforcement learning for real-world manipulation is constrained by interaction costs and safety risks. To reduce reliance on real-robot trial and error, some research focuses on learning action-conditioned world models and optimizing policies through imagined interactions~\citep{ gigaworldteam2025gigaworld0worldmodelsdata, li2025roboticworldmodelneural, zhang2026practicalworldmodelbasedreinforcement, jiang2026world4rldiffusionworldmodels}. WoVR~\citep{jiang2026wovrworldmodelsreliable} addresses prediction errors and their accumulation to improve the reliability of model-generated rollouts for VLA reinforcement fine-tuning 
, while PlayWorld~\citep{yin2026playworldlearningrobotworld} and RISE~\citep{yang2026riseselfimprovingrobotpolicy} learn world models from autonomous robot interactions and transfers policies trained through model-based reinforcement learning to real hardware. 
Although world models can effectively capture short-term dynamics, long-horizon predictions still exhibit visual distortions and action–state inconsistencies. Other studies adopt an explicit Real2Sim2Real method, combining real-scene reconstruction with physics simulation to provide interactive environments for policy optimization~\citep{li2025pinwmlearningphysicsinformedworld, wang2026phys2realfusingvlmpriors}. GSWorld~\citep{jiang2025gsworldclosedloopphotorealisticsimulation} integrates Gaussian-on-Mesh assets with physics engines to support visually faithful policy training, while 
RialTo~\citep{torne2024reconcilingrealitysimulationrealtosimtoreal} constructs digital twins of real-world environments and uses reinforcement learning in simulation to improve policy robustness.
However, constructing environments requires coordinating scene reconstruction, physical asset preparation, and control alignment, making it complex and costly to reproduce. AtomicRL introduces an automated Real2Sim2Real pipeline that uses only a single demonstration to reconstruct a real-world manipulation scene as an interactive simulation environment for long-horizon online RL. The trained policies are then transferred to physical robots for evaluation.

%% file: sec/3_methodology.tex
\begin{figure}[t]
    \centering
    \includegraphics[width=1.0\textwidth]{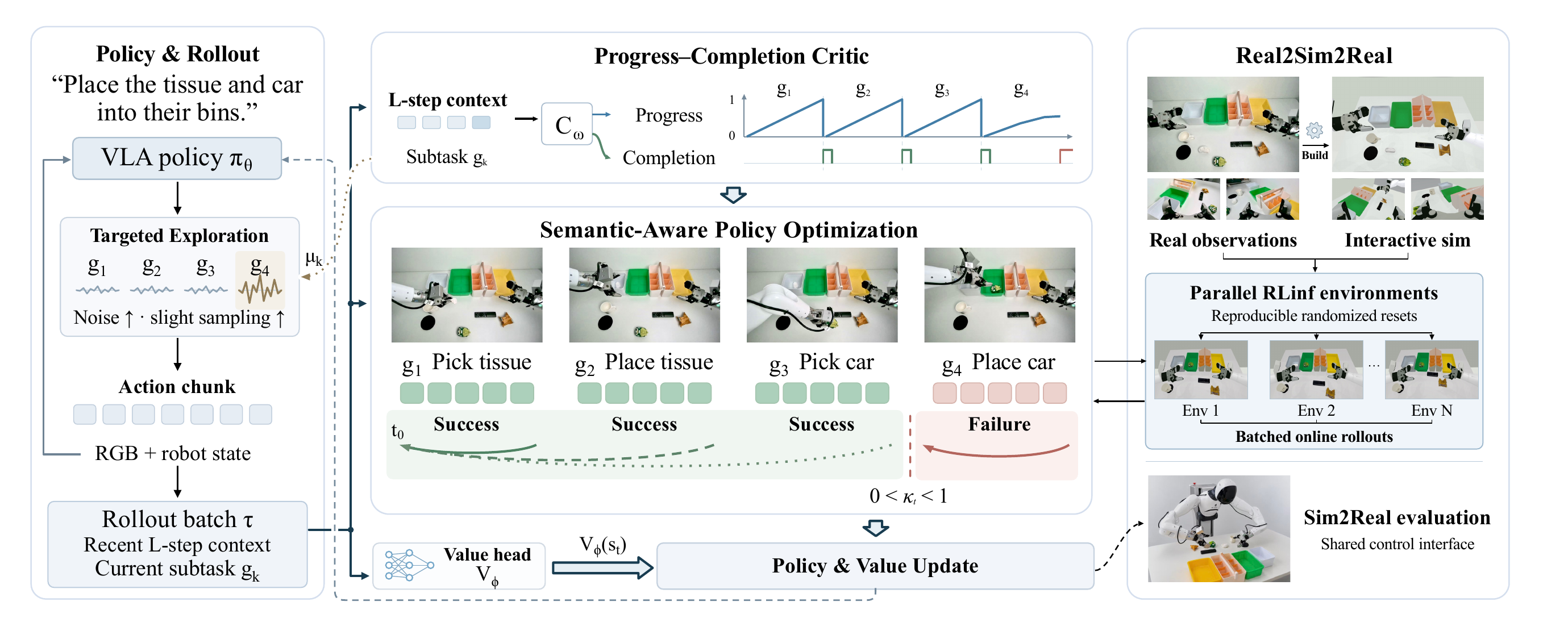}
    \caption{\textbf{Overall pipeline of AtomicRL.} \textbf{(a) Progress–Completion Critic:} PCC identifies atomic task boundaries and estimates atomic task progress from interaction histories. \textbf{(b) Semantic-Aware Policy Optimization:} SAPO uses these signals to regulate cross-subtask advantage propagation based on subtask execution outcomes. \textbf{(c) Bottleneck-Adaptive Targeted Exploration:} BATE adapts task sampling and local exploration according to subtask mastery, focusing training on insufficiently mastered bottleneck subtasks. \textbf{(d) Real2Sim2Real workflow:} A complementary Real2Sim2Real pipeline enables parallel simulation-based training and transfer to real-world  robots.}
    \label{pipline} 
    \vspace{-12pt} 
\end{figure}

\section{Method}
\subsection{Overview}

As illustrated in Figure~\ref{pipline}, AtomicRL integrates semantic-aware credit assignment and bottleneck-adaptive exploration for long-horizon robotic manipulation. Specifically, SAPO modulates credit propagation across atomic subtasks (Section~\ref{sec:sapo}), while bottleneck-adaptive targeted exploration adjusts full-task trajectory sampling and local exploration incentives (Section~\ref{sec:exploration}). Both components are supported by a Progress--Completion Critic (PCC), which estimates subtask progress and identifies completion boundaries (Section~\ref{sec:phc}). For real-world tasks, we further introduce a Real2Sim2Real construction workflow to support simulation training and subsequent transfer evaluation (Section~\ref{sec:real2sim}).

\subsubsection{Problem Formulation}
\label{sec:problem}

We consider a language-conditioned partially observable Markov decision process for long-horizon robotic manipulation. At each decision step $t$, the robot receives an observation $o_t$, which may include third-person images, wrist-camera images, and proprioceptive states, and outputs an action chunk $a_t=(a_{t,1},\ldots,a_{t,H_a})$, where $H_a$ denotes the chunk length. A long-horizon instruction is decomposed into an ordered sequence of $K$ atomic subtasks, $G=(g_1,\ldots,g_K)$. We denote the active subtask index by $z_t\in{1,\ldots,K}$. For nonterminal transitions, the semantic-boundary indicator is defined as $b_t=\mathbb{I}[z_{t+1}\neq z_t]$, indicating whether execution of $a_t$ leads to a subtask transition. Our objective is to maximize full-task success under a fixed environment-interaction budget by improving credit assignment and directing exploration toward bottleneck subtasks.

\subsection{Semantic-Aware Policy Optimization}
\label{sec:sapo}

SAPO addresses temporal credit assignment in long-horizon tasks. It leverages semantic boundaries to selectively attenuate the contribution of TD residuals from failed subtasks to the advantage estimates of preceding actions, thereby regulating cross-subtask credit propagation.

Let $r_t$ denote the chunk-level reward and $d_t$ indicate environment termination after executing $a_t$. The one-step TD residual is
\begin{equation}
\delta_t
=
r_t
+
\gamma(1-d_t)V_{\phi}(s_{t+1})
-
V_{\phi}(s_t),
\label{eq:td_residual}
\end{equation}
where $\gamma$ is the discount factor and $s_t$ denotes the observation- and language-conditioned input representation available to the value function, rather than a privileged simulator state. For an episode containing $T$ chunk-level transitions indexed by $t=1,\ldots,T$, the standard GAE estimate used by PPO is
\begin{equation}
\hat{A}^{\mathrm{PPO}}_t
=
\sum_{\ell=0}^{T-t}
(\gamma\lambda)^\ell
\delta_{t+\ell},
\label{eq:standard_gae_simplified}
\end{equation}
where $\lambda\in[0,1]$ controls trace decay. This expression is evaluated within a single episode and does not span environment resets.

SAPO distinguishes semantic segmentation from environment termination and uses atomic subtask outcomes to regulate cross-boundary credit propagation. Specifically, PCC (Section~\ref{sec:phc}) identifies semantic boundaries $b_t$ between atomic subtasks. Keeping the one-step TD residual unchanged, SAPO applies an outcome-aware soft gate to the recursive GAE trace:
\begin{equation}
\hat{A}^{\mathrm{SAPO}}_t
=
\sum_{\ell=0}^{T-t}
(\gamma\lambda)^\ell
\left[
\prod_{j=0}^{\ell-1}
\left(
1-(1-\kappa_{t+j})b_{t+j}
\right)
\right]
\delta_{t+\ell},
\qquad \kappa_t\in[0,1].
\label{eq:sapo_gae}
\end{equation}
Within a subtask, $b_t=0$, and the trace follows the standard $\gamma\lambda$ decay. At a semantic boundary, $\kappa_t=1$ when the subsequent subtask succeeds, preserving ordinary propagation; when it fails, we set $0<\kappa_t<1$ to attenuate propagation across the boundary. Thus, when subtasks $1{:}k$ succeed but subtask $k+1$ fails, residuals from the failed segment propagate normally within that segment and are additionally attenuated when entering the completed prefix. Residuals within the completed prefix retain their standard trace weights. This design preserves within-task credit propagation and connections between successfully completed subtasks while reducing the direct negative influence of downstream failures on earlier successful behavior. We use the resulting advantages for updates and construct the value targets as $V_\phi(s_t)+\hat{A}^{\mathrm{SAPO}}_t$ before advantage normalization. By making credit propagation sensitive to semantic boundaries and local outcomes, SAPO improves credit assignment and interaction-data efficiency in long-horizon tasks.

\subsection{Bottleneck-Adaptive Targeted Exploration}
\label{sec:exploration}

During the early stages of RL, broad exploration provides informative samples for most atomic subtasks. As training approaches saturation, failures may concentrate in a few bottleneck subtasks, making persistent aggressive exploration less effective and potentially disruptive to acquired skills. BATE adapts task sampling and local exploration incentives according to atomic subtask mastery.

We estimate the mastery level of each atomic-task category using an exponential moving average of its completion rate:

\begin{equation}
\mu_k^{(n)}
=
\beta\mu_k^{(n-1)}
+
(1-\beta)\hat{s}_k^{(n)},
\qquad \beta\in[0,1),
\label{eq:subtask_mastery}
\end{equation}
where $n$ indexes statistics updates and $\hat{s}_k^{(n)}$ is the fraction of successful outcomes among subtask attempts whose outcomes are resolved during the update. 

The competence gap, $1-\mu_k$, guides both full-task sampling and local exploration. 

\begin{equation}
w_i
=
1+\rho
\left[
\max_{k\in\mathcal{K}_i}(1-\mu_k)
\right]^\eta,
\qquad
P_{\mathrm{task}}(i)
=
\frac{w_i}{\sum_j w_j},
\label{eq:bottleneck_task_sampling}
\end{equation}
where $\mathcal K_i$ contains the subtasks of full task $i$, $\eta>0$ controls sensitivity to the competence gap, and $\rho\geq0$ limits sampling redistribution. The baseline weight of $1$ preserves coverage of mastered tasks to monitor forgetting.

Within each trajectory, mastery-dependent entropy regularization adjusts local exploration:

\begin{equation}
\mathcal{L}_{\mathrm{explore}}
=
-\alpha_H
\mathbb{E}_t
\left[
f_{\mathrm{exp}}\!\left(1-\mu_{z_t}^{(n)}\right)
\mathcal{H}_{\theta,t}
\right],
\label{eq:targeted_exploration}
\end{equation}
where $\alpha_H$ is the base entropy coefficient, $\mathcal H_{\theta,t}$ denotes policy entropy, and $f_{\mathrm{exp}}$ is a bounded exploration weight for the active subtask $z_t$. The exploration strength is thus increased for poorly mastered subtasks and reduced as subtask mastery improves.

\subsection{Progress--Completion Critic}
\label{sec:phc}

To identify atomic-task boundaries $b_t$, we introduce a lightweight Progress--Completion Critic (PCC), denoted by $C_\omega$. PCC estimates within-task progress and completion from the recent $L$ observation--action pairs, providing semantic signals for SAPO and BATE and support the construction of progress-based shaping rewards (PBSR).

Conditioned on the current atomic task, PCC evaluates its status after executing $a_t$ and outputs $B-1$ ordinal progress logits $u_{t,j}^{\mathrm{prog}}$ and a completion logit $u_t^{\mathrm{comp}}$, where $B$ is the number of progress levels. We decode progress as
\begin{equation}
\hat{p}_t
=
\frac{1}{B-1}
\sum_{j=1}^{B-1}
\mathbf{1}\!\left[
\sigma\!\left(u_{t,j}^{\mathrm{prog}}\right)
>
\tau_{\mathrm{ord}}
\right],
\qquad
\hat{q}_t
=
\sigma\!\left(u_t^{\mathrm{comp}}\right),
\label{eq:phc_predictions}
\end{equation}
where $\sigma$ denotes the sigmoid function and $\tau_{\mathrm{ord}}=0.5$ is the ordinal decoding threshold. An atomic task is considered complete when $\hat{q}_t\geq\tau_s$ and $\hat{p}_t\geq\tau_p$ for $N$ consecutive decision steps. Upon confirming a non-terminal boundary, we advance the atomic-task index as $z_{t+1}=z_t+1$ and reset the tracked within-task progress while retaining the physical observation context.
PCC is trained offline on demonstration data using a cumulative ordinal loss together with a class-balanced completion loss:
\begin{equation}
\mathcal{L}_C
=
\lambda_p\mathcal{L}_{\mathrm{ord}}
+
\lambda_c\mathcal{L}_{\mathrm{comp}},
\label{eq:phc_loss}
\end{equation}
where $\mathcal{L}_{\mathrm{ord}}$ applies binary cross-entropy to cumulative progress targets, $\mathcal{L}_{\mathrm{comp}}$ is a class-balanced binary cross-entropy loss for completion prediction. The coefficients $\lambda_p$ and $\lambda_c$ control the relative contributions of the two objectives.
The progress estimates can optionally be used to construct
progress-based shaping rewards (PBSR), yielding the training reward
\begin{equation}
r_t
=
r_t^{\mathrm{env}}
+
\alpha_n r_t^{\mathrm{PBSR}},
\label{eq:training_reward}
\end{equation}
where $r_t^{\mathrm{env}}$ is the sparse task reward, $r_t^{\mathrm{PBSR}}$ is the progress-based shaping reward.

\subsection{Real2Sim2Real Pipeline for Reinforcement Learning}
\label{sec:real2sim}
We develop an automated Real2Sim2Real reinforcement learning pipeline that unifies interactive environment construction, policy optimization in simulation, and deployment and evaluation on physical robots. Given a single multimodal real-world demonstration \(\tau_{\mathrm{real}}\), metadata \(m\) specifying the robot model and supplementary object observations \(\mathcal D_{\mathrm{aux}}\) where needed, the pipeline constructs a simulation environment aligned with the physical system, a shared observation–action interface specification, and a set of reproducible randomized initial layouts: 
\begin{equation}
(\mathcal E_{\mathrm{sim}},\mathcal C,\mathcal S_0)=\mathcal B(\tau_{\mathrm{real}},m,\mathcal D_{\mathrm{aux}}),
\end{equation}
where \(\mathcal E_{\mathrm{sim}}\) is the reconstructed interactive simulation environment, \(\mathcal C\) specifies the shared observation–action interface, \(\mathcal S_0\) is a set of reproducible randomized initial layouts, and \(\mathcal B\) denotes the pipeline workflow. 
The generated environment is then integrated into the RLinf~\citep{yu2025rlinfflexibleefficientlargescale} distributed training framework, where SAPO and targeted exploration optimize the initial policy to obtain \(\theta_{\mathrm{RL}}=\operatorname{Optimize}(\theta_0;\mathcal E_{\mathrm{sim}},\mathcal C,\mathcal S_0)\). Following pre-deployment validation, the optimized policy is deployed on the physical robot through the same interface specification \(\mathcal C\) for task evaluation. 

%% file: sec/4_experiment.tex
\section{Experiment}
\label{sec:experiment}

\subsection{Experiments Setup}

\mypara{Simulation Benchmarks.}
We evaluate AtomicRL on two robotic manipulation benchmarks: LIBERO~\cite{liu2023liberobenchmarkingknowledgetransfer} and CALVIN~\cite{mees2022calvinbenchmarklanguageconditionedpolicy}. We use the LIBERO-LONG suite(10 long-horizon tasks), and for CALVIN, we follow the ABC-D protocol, training on environments A, B, and C and evaluating on the held-out environment D. All experiments are implemented within the RLinf training framework~\cite{yu2025rlinfflexibleefficientlargescale}. On LIBERO-LONG, we compare AtomicRL against PPO with two policy backbones, $\pi_{0.5}$~\cite{intelligence2025pi05visionlanguageactionmodelopenworld} and OpenVLA-OFT~\cite{kim2025finetuningvisionlanguageactionmodelsoptimizing}; on CALVIN, we conduct additional evaluations with $\pi_{0.5}$. Each training run consists of 200 optimization steps. 

\mypara{Real World Robot.}
We conduct real-world experiments using an AgiBot G1 robot across three long-horizon tabletop manipulation tasks, each consisting of 4 to 6 atomic subtasks. For each task, we collect 100 expert demonstrations for SFT. The three tasks are:

\begin{itemize}[leftmargin=1.5em, itemsep=0pt, topsep=2pt]
\item \textbf{Bread Placement(BP):} Place all pieces of bread on the table onto the white plate;
\item \textbf{Waste Disposal(WD):} Place the tissue and the car into the trash bin;
\item \textbf{Tabletop Clean(TC):} Place the remote control into the storage box and the snacks into the box.
\end{itemize}

\mypara{Evaluation metrics.} For LIBERO, we report the mean success rate across all tasks, with 50 trials per task. For CALVIN, we report the success rates for completing one to five consecutive tasks and the average completed sequence length over 1,000 evaluation trials. These metrics measure cumulative sequence completion rather than independent subtask performance. For real-robot evaluation, each task is tested over 25 trials per method, and we report the mean success rate across the three tasks.

\begin{table}[t]
\centering
\small
\setlength{\tabcolsep}{8pt}
\renewcommand{\arraystretch}{1.12}

\caption{\textbf{Evaluation on LIBERO-LONG (\%).}
$\Delta_{\mathrm{PPO}}$ denotes the improvement over PPO.}
\label{tab:libero_main}

\begin{tabular}{
    l
    l
    c
    c
}
\toprule

\textbf{Backbone}
& \textbf{Method}
& \textbf{SR (\%) $\uparrow$}
& \textbf{$\Delta_{\mathrm{PPO}}$ (pp)} \\

\midrule

\multirow{3}{*}{$\pi_{0.5}$}
& SFT
& 89.0
& -- \\

& \quad + PPO
& 95.4
& -- \\

\rowcolor{mygray}
& \quad \textbf{+ AtomicRL (Ours)}
& \textbf{99.0}
& \cellcolor{myblue}\textbf{+3.6} \\

\midrule

\multirow{3}{*}{OpenVLA-OFT}
& SFT
& 86.5
& -- \\

& \quad + PPO
& 94.8
& -- \\

\rowcolor{mygray}
& \quad \textbf{+ AtomicRL (Ours)}
& \textbf{98.8}
& \cellcolor{myblue}\textbf{+4.0} \\

\bottomrule
\end{tabular}

\vspace{-6pt}
\end{table}

\begin{table*}[t]
\centering
\small
\setlength{\tabcolsep}{7pt}
\renewcommand{\arraystretch}{1.12}

\caption{\textbf{Long-horizon Robotic Manipulation Evaluation on CALVIN Benchmark (\%).}}
\label{tab:calvin_main}

\begin{tabular}{
    l
    c
    cccc
    >{\columncolor{myblue}}c
    c
}
\toprule

\multirow{2}{*}{\textbf{Method}}
& \multirow{2}{*}{\textbf{Task}}
& \multicolumn{5}{c}{\textbf{Tasks Completed in a Row}}
& \multirow{2}{*}{\textbf{Avg.\ Len $\uparrow$}} \\

\cmidrule(lr){3-7}

& & 1 & 2 & 3 & 4 & 5 & \\

\midrule

$\pi_{0.5}$ SFT
& ABC$\rightarrow$D
& 92.8
& 83.5
& 73.2
& 61.5
& 45.3
& 3.56 \\

\quad + PPO
& ABC$\rightarrow$D
& 99.2
& 95.7
& 89.8
& 81.2
& 70.6
& 4.36 \\

\midrule

\rowcolor{mygray}
\quad \textbf{+ AtomicRL (Ours)}
& ABC$\rightarrow$D
& \textbf{99.4}
& \textbf{97.0}
& \textbf{92.6}
& \textbf{87.3}
& \cellcolor{myblue}\textbf{80.1}
& \textbf{4.56} \\

\bottomrule
\end{tabular}

\vspace{-15pt}
\end{table*}

\subsection{Results on Simulation}
\mypara{Results on LIBERO.}
As shown in Tab.~\ref{tab:libero_main}, AtomicRL achieves success rates of $98.8\%$ with OpenVLA-OFT and $99.0\%$ with $\pi_{0.5}$, exceeding their corresponding PPO baselines by $4.0\%$ and $3.6\%$. The consistent gains across both backbones support the applicability of AtomicRL to different VLA architectures. Both policies reach near-perfect success rates after 200 optimization iterations, demonstrating further improvements even when PPO already performs well. 
This efficiency can be attributed to AtomicRL's effective use of training data: SAPO preserves useful learning signals from completed subtasks in failed trajectories, while BATE concentrates exploration on bottleneck atomic subtasks. Together, these mechanisms enable faster convergence and higher final performance. 


\mypara{Results on CALVIN.} As shown in Tab.~\ref{tab:calvin_main}, AtomicRL improves the average completed sequence length from $3.56$ with SFT to $4.56$, compared with $4.36$ for PPO. The success rate for completing all five tasks increases from $70.6\%$ with PPO to $80.1\%$, a gain of $9.5\%$. The larger gains on longer sequences indicate more reliable execution across the full task chain, demonstrating the effectiveness of AtomicRL for long-horizon reinforcement learning.

\subsection{Results on Real-world Robot}

\begin{figure}[h]
    \centering

    \begin{subfigure}[t]{0.32\linewidth}
        \centering
        \includegraphics[width=\linewidth]{
            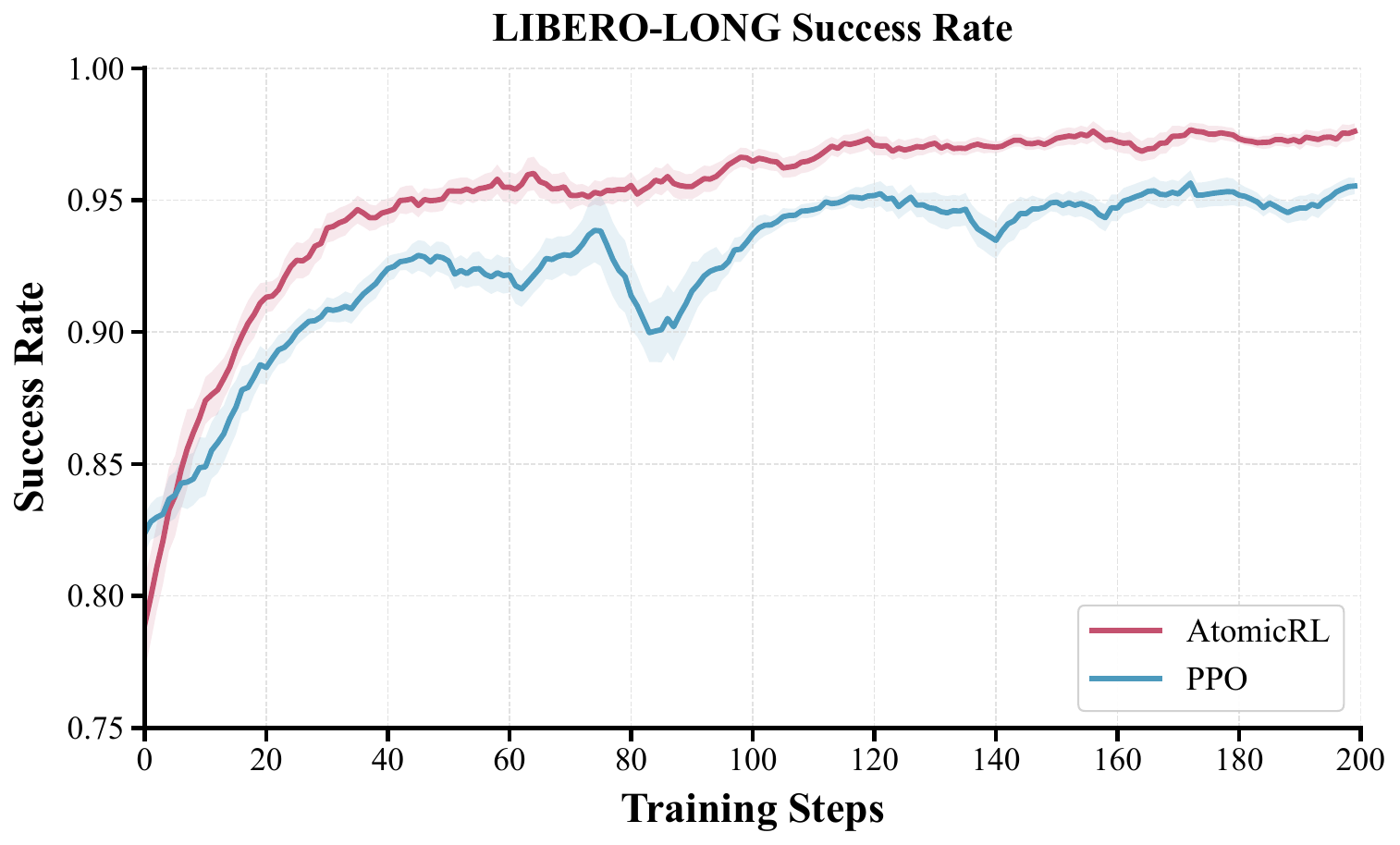
        }
        \caption{LIBERO-LONG.}
        \label{fig:libero10-success-once}
    \end{subfigure}
    \hfill
    \begin{subfigure}[t]{0.32\linewidth}
        \centering
        \includegraphics[width=\linewidth]{
            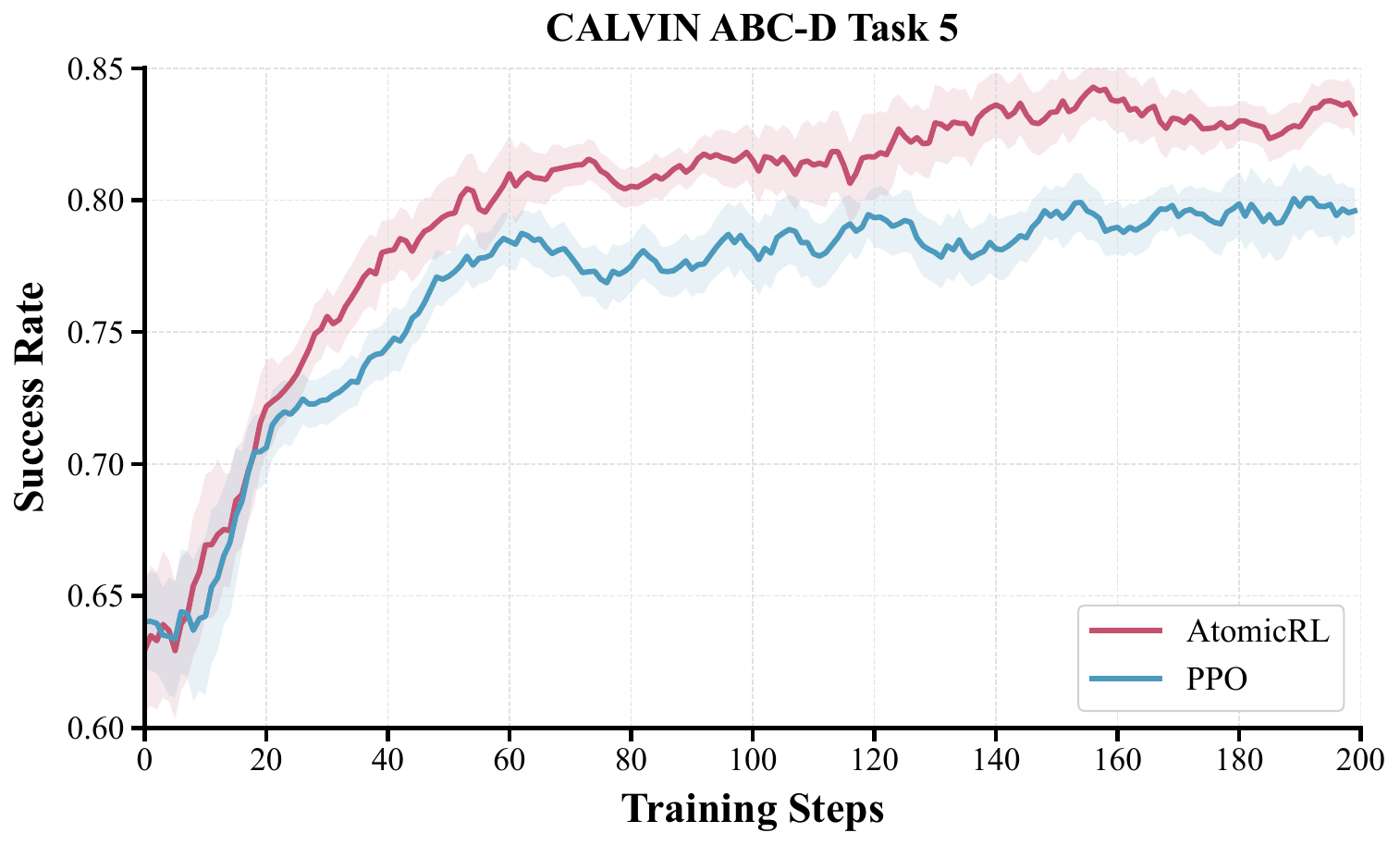
        }
        \caption{CALVIN ABC-D.}
        \label{fig:calvin-success-once}
    \end{subfigure}
    \hfill
    \begin{subfigure}[t]{0.32\linewidth}
        \centering
        \includegraphics[width=\linewidth]{
            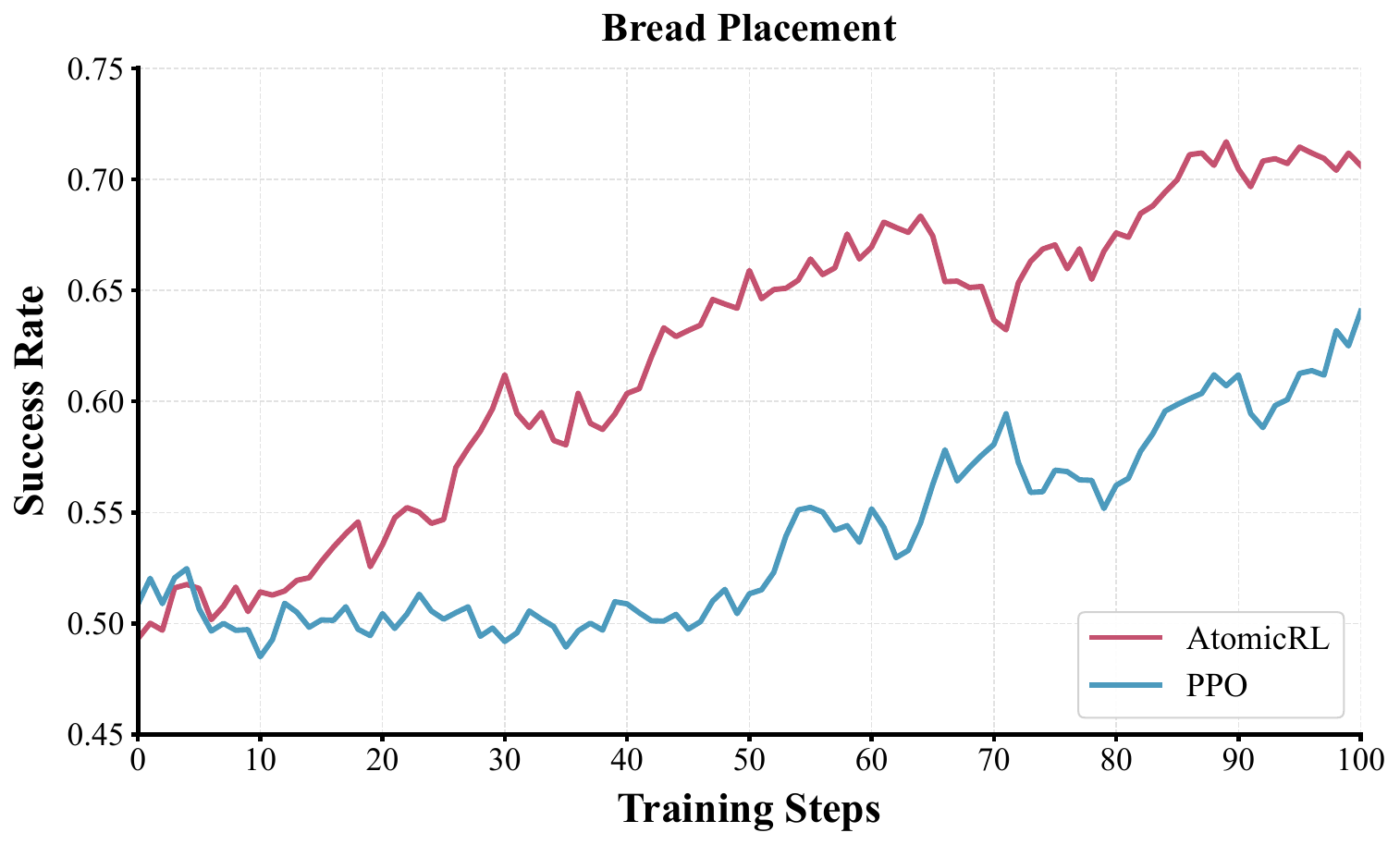
        }
        \caption{Bread placement.}
        \label{fig:bread}
    \end{subfigure}

    \caption{
         Training curves on LIBERO-LONG and CALVIN ABC-D and Real-World(BP).
    }
    \label{fig:training-curves}
\end{figure}

We conduct online reinforcement learning in the constructed real-to-sim environments and evaluate both the policy improvement after RL and its sim-to-real transfer performance.
\begin{wraptable}{r}{0.48\textwidth}
    \centering
    \small
    \vspace{-10pt}
    \setlength{\tabcolsep}{5pt}
    \renewcommand{\arraystretch}{1.12}

    \caption{\textbf{Real-world robot results(\%).}}
    \label{tab:real_robot}

    \begin{tabular}{lccc>{\columncolor{myblue}}c}
        \toprule
        \textbf{Method}
        & \textbf{BP}
        & \textbf{WD}
        & \textbf{TC}
        & \textbf{Avg.\ SR $\uparrow$} \\
        \midrule

        $\pi_{0.5}$ SFT
        & 60 & 52 & 32 & 48.0 \\

        \quad + PPO
        & 68 & 64 & 52 & 61.3 \\
        \midrule

        \rowcolor{mygray}
        \quad \textbf{+ AtomicRL}
        & \textbf{76}
        & \textbf{72}
        & \textbf{60}
        & \cellcolor{myblue}\textbf{69.3} \\
        \bottomrule
    \end{tabular}
    \vspace{-10pt}
\end{wraptable}
\mypara{Performance on real-world robot.} As shown in Tab.~\ref{tab:real_robot}, AtomicRL improves over SFT by 21.3\% on average across three long-horizon tasks and over PPO by 8.0\%. As shown in Fig.~\ref{fig:training-curves}, SAPO preserves useful learning signals from successfully completed subtasks in unsuccessful trajectories, enabling more appropriate credit assignment for suboptimal trajectories. Consequently, AtomicRL improves faster in early training, especially on the longer-horizon Bread Placement task. These results demonstrate the effectiveness of AtomicRL for long-horizon manipulation.

\begin{wrapfigure}{r}{0.39\textwidth}
    \vspace{-10pt}
    \centering
    \includegraphics[width=\linewidth]{
        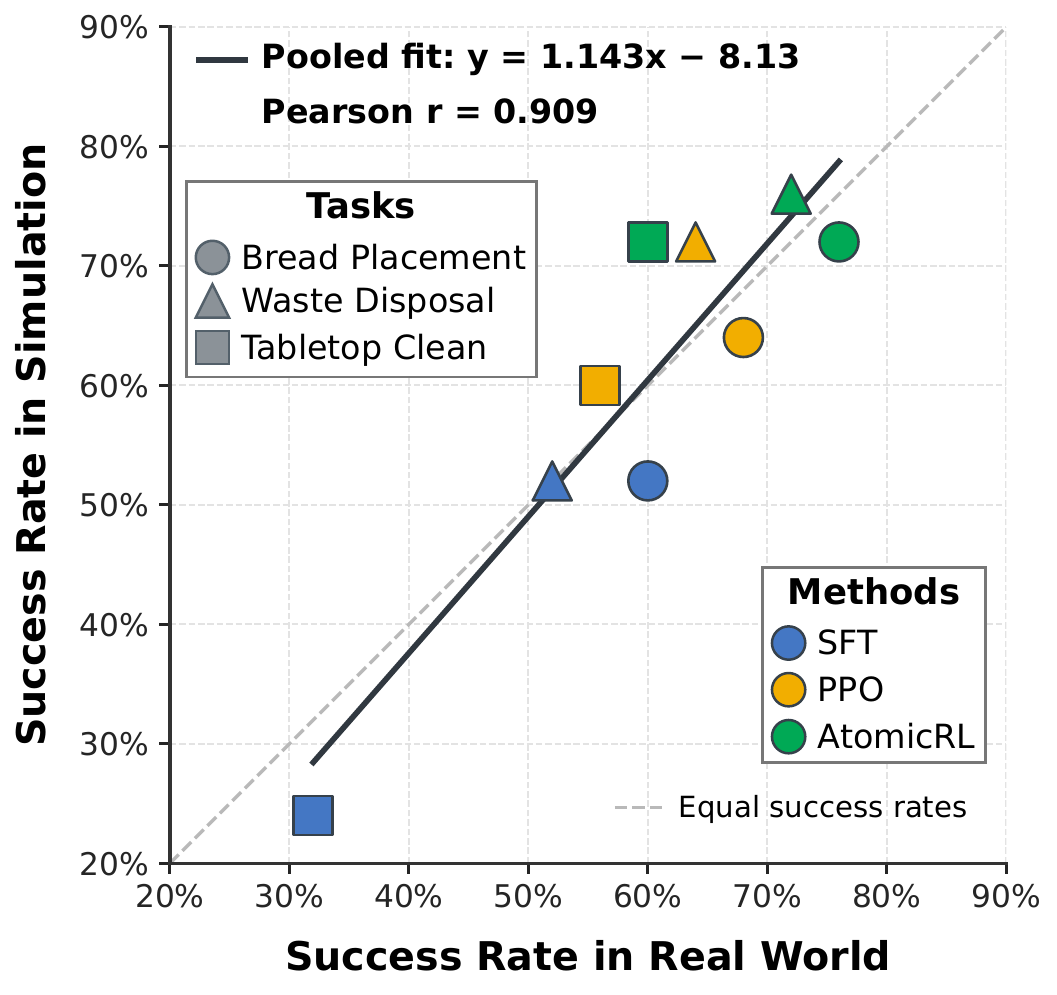
    }
    \vspace{-15pt}
    \caption{\textbf{Sim2Real Correlation.}}
    \label{fig:real_sim_transfer}
    \vspace{-25pt}
\end{wrapfigure}

\begin{figure}[h]
    \centering
    \vspace{-5pt}
    \includegraphics[width=1.0\textwidth]{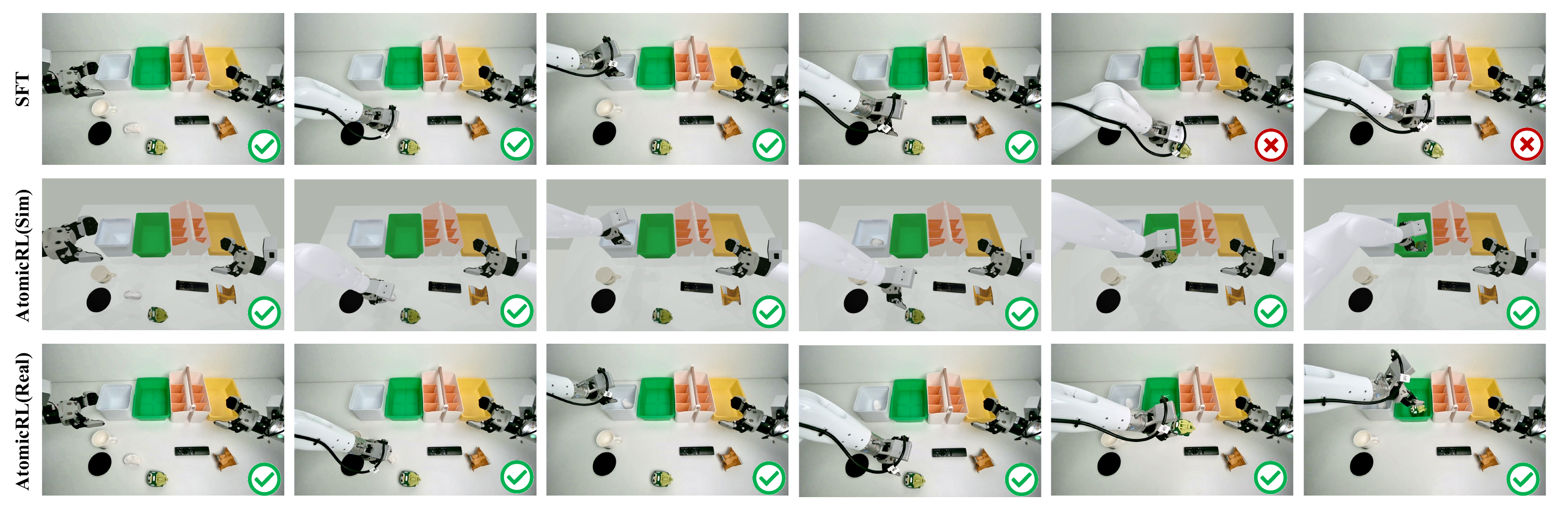}
    \caption{Demonstrations show the execution process of SFT and AtomicRL.}
    \label{vis} 
    \vspace{-12pt} 
\end{figure}

\mypara{Simulation–reality correspondence.} 
Fig.~\ref{fig:real_sim_transfer} compares the success rates between simulation and the real robot across nine combinations of three tasks and three methods. The aggregated Pearson correlation coefficient is \(r=0.909\), indicating a strong positive correlation across the evaluated settings. 
As shown in Fig.~\ref{vis}, the SFT policy fails after completing only part of the task, whereas the AtomicRL-optimized policy successfully completes the illustrated sequence in both simulation and the real world. The qualitative consistency in observations and execution outcomes between the two environments indicates that the reconstructed environment preserves the key visual and interaction structures relevant to the task, supporting effective policy optimization. Combined with the Real2Sim2Real deployment pipeline, AtomicRL shifts the otherwise costly and operationally complex online RL process from physical robots to simulation environments that support fast and parallel interactions. This reduces real-world interaction and reset costs while providing a practical approach for optimizing and deploying long-horizon manipulation policies.

\subsection{Ablation Study}
We conduct policy-learning ablations on LIBERO-LONG using a weaker SFT checkpoint trained with fewer demonstrations than those used in the main evaluations. This few-shot initialization allows us to examine the contribution of each component under a less proficient initial policy.

\mypara{Reward shaping and SAPO.} To investigate the effect of reward granularity, we use the PCC module to construct process-level dense rewards and subtask-level rewards, and compare them under both PPO and SAPO. As shown in Tab.~\ref{tab:component_ablation} (a), denser rewards do not always yield better performance: under SAPO, relatively sparse subtask-level rewards outperform process-level dense rewards. When the policy success rate is low, dense rewards provide stronger exploration guidance and accelerate early learning. As the policy improves, however, their benefits diminish, while persistent auxiliary rewards may dilute the terminal success signal and shift optimization toward local gains. Motivated by this observation, we use linear annealing to gradually reduce the PBSR weight, providing stronger dense-reward guidance early in training and emphasizing the sparse terminal reward later. This balances early exploration with late-stage task alignment.

\begin{table}[h]
\centering
\small
\caption{\textbf{Ablations on LIBERO-LONG.}
(a) Reward shaping with and without SAPO. (b) Component contributions. All results are success rates (\%).}
\label{tab:component_ablation}

\setlength{\tabcolsep}{5pt}
\renewcommand{\arraystretch}{1.15}

\begin{minipage}[t]{0.48\textwidth}
    \centering
    \textbf{(a) Reward shaping and SAPO}
    \par\vspace{6pt}

    \begin{tabularx}{\linewidth}{
        >{\raggedright\arraybackslash}X
        >{\centering\arraybackslash}X
        >{\centering\arraybackslash}X
    }
        \toprule
        \textbf{Reward}
        & \textbf{PPO}
        & \textbf{SAPO} \\
        \midrule

        0/1  & 78.2 & 87.8 \\
        SubR & 80.2 & 89.4 \\
        PBSR & 81.8 & 89.0 \\

        \rowcolor{mygray}
        \textbf{Ours} & -- & \textbf{89.6} \\

        \bottomrule
    \end{tabularx}
\end{minipage}
\hfill
\begin{minipage}[t]{0.48\textwidth}
    \centering
    \textbf{(b) Full-method ablation}
    \par\vspace{6pt}

    \begin{tabularx}{\linewidth}{
        >{\centering\arraybackslash}X
        >{\centering\arraybackslash}X
        >{\centering\arraybackslash}X
        >{\centering\arraybackslash}X
    }
        \toprule
        \textbf{SAPO}
        & \textbf{PBSR}
        & \textbf{BATE}
        & \textbf{SR (\%)} \\
        \midrule

        $\times$
        & $\times$
        & $\times$
        & 78.2 \\

        $\checkmark$
        & $\times$
        & $\times$
        & 87.8 \\

        $\checkmark$
        & $\checkmark$
        & $\times$
        & 89.6 \\

        \rowcolor{mygray}
        $\checkmark$
        & $\checkmark$
        & $\checkmark$
        & \textbf{91.2} \\
        \bottomrule
    \end{tabularx}
\end{minipage}
\vspace{-20pt}

\end{table}

\mypara{Component analysis.} Table~\ref{tab:component_ablation} (b) presents the incremental effects of successively introducing SAPO, annealed PBSR, and BATE. Starting from a PPO success rate of 78.2\%, adding SAPO increases the success rate to 87.8\% (+9.6 pp); introducing annealed PBSR further improves it to 89.6\% (+1.8 pp); and finally incorporating BATE raises it to 91.2\% (+1.6 pp), yielding a cumulative improvement of 13.0 percentage points over PPO. SAPO provides the largest performance gain, while annealed reward shaping and bottleneck-adaptive exploration further improve the policy. Combined with the results in Table~\ref{tab:component_ablation} (a), where SAPO consistently improves performance across different reward settings, these findings suggest that its gains do not simply arise from denser reward feedback. Together, these results support the complementarity of the proposed components and their ability to further unlock the potential of RL for learning long-horizon tasks.


\mypara{PCC prediction quality and computational cost.} We train PCC on 150 trajectories containing both successful and failed executions and evaluate it on a test set of 500 rollout trajectories. The reference annotations for training and evaluation are automatically generated by a VLM and subsequently refined through manual correction. As shown in Tab.~\ref{tab:pcc_performance}, PCC achieves 95.92\% precision and 89.20\% recall, with an empirical inference latency of 42.4 ms. 
\begin{wraptable}{r}{0.45\textwidth}
    \vspace{-6pt}
    \centering

    \caption{\textbf{PCC performance and efficiency.}}
    \label{tab:pcc_performance}

    \footnotesize
    \setlength{\tabcolsep}{3pt}
    \renewcommand{\arraystretch}{1.12}

    \begin{tabularx}{\linewidth}{
        @{}l
        >{\raggedright\arraybackslash}X
        r@{}
    }
        \toprule
        \textbf{Category}
        & \textbf{Metric}
        & \textbf{Value} \\
        \midrule

        \multirow{3}{*}{\shortstack[l]{Prediction\\quality}}
        & Accuracy
        & 95.92\% \\

        & False detection rate
        & 4.08\% \\

        & Recall
        & 89.20\% \\

        \midrule

        \multirow{2}{*}{Efficiency}
        & Trainable Parameters
        & 28.8 M \\

        & Inference latency
        & 42.4 ms \\

        \bottomrule
    \end{tabularx}

    \vspace{-6pt}
\end{wraptable}
These results show that PCC can estimate subtask progress and completion status with high precision while maintaining low inference latency. As a core component of AtomicRL, PCC converts the semantic task structure in trajectories into progress and boundary information for policy optimization, supporting SAPO in regulating cross-subtask credit propagation and BATE in estimating subtask mastery. Its prediction accuracy and low inference overhead demonstrate that this lightweight design provides a practical perception and evaluation foundation for RL based on atomic task structure.

%% file: sec/5_conclusion.tex
\vspace{-3mm}
\section{Conclusion}
\label{sec:conclusion}
In this work, we introduce AtomicRL, a VLA RL post-training framework for long-horizon robotic manipulation centered on atomic task structure. To address temporal  credit interference and imbalanced atomic subtask mastery, AtomicRL uses PCC to estimate task progress and completion boundaries, SAPO to regulate the cross-boundary propagation of TD residuals based on subtask execution outcomes, and BATE to shift task sampling and local exploration toward insufficiently mastered bottleneck subtasks. We further develop a Real2Sim2Real pipeline that connects real-task reconstruction, simulation-based RL, and real-robot deployment, enabling policy improvement in simulation environments with fast parallel interactions and reproducible resets. We validate AtomicRL in both simulation and real-world settings, and the results show that it converges faster to a higher performance level, demonstrating its potential for RL-based long-horizon robotic manipulation.

%% file: sec/X_suppl.tex
\appendix
\newpage

\section{Appendix/supplemental material}

\subsection{Future Work and Limitations}
AtomicRL improves temporal credit assignment across atomic tasks through SAPO and concentrates exploration resources on insufficiently mastered bottleneck subtasks through BATE. Although the experimental results demonstrate its effectiveness in long-horizon robotic manipulation, the current approach still has several limitations.

\begin{itemize}[leftmargin=1.5em, itemsep=0pt, topsep=2pt]
\item AtomicRL improves the utilization of interaction data through more effective credit assignment and exploration-resource allocation, but does not fundamentally address the overall efficiency of online RL training. Even with open-source RL frameworks such as RLinf~\citep{yu2025rlinfflexibleefficientlargescale} and SERL~\citep{luo2025serlsoftwaresuitesampleefficient}, online RL still requires substantial environment interaction and training time, particularly in real-robot settings where interaction and exploration costs remain high. Reducing the interaction required for effective policy learning, improving online RL training efficiency, and lowering the cost of real-robot training therefore remain important directions for future work.
\item Our experiments primarily focus on long-horizon tabletop manipulation tasks and initialize RL from SFT policies. The effectiveness of AtomicRL on more open and dynamic tasks requiring stronger embodiment generalization remains to be further evaluated. Recent Agentic RL approaches, such as HarnessVLA~\citep{zhang2026harnessvlasteeringfrozen} and AtomBridge~\citep{pang2026atombridgeagenticvlainference} have shown potential for open-world tasks and embodiment generalization, providing an alternative direction for long-horizon robotic manipulation. A promising direction is to transfer AtomicRL's use and optimization of atomic task structure to Agentic RL frameworks, enabling robots to dynamically form and adapt atomic task structures in more open environments and potentially improving the generalization and training efficiency of long-horizon manipulation.
\end{itemize}

\subsection{Progress-Based Shaping Rewards}
\label{app:pbsr}

We optionally supplement the sparse task reward with progress-based shaping rewards (PBSR), using the atomic-task index and discrete progress estimates from PCC.

\paragraph{Progress potential.}
Let $K$ denote the number of atomic tasks and
$c_t=(B-1)\hat{p}_t \in \{0,\ldots,B-1\}$ denote
the decoded progress level.
After completion handling, let $z_{t+1}$ be the
active atomic-task index and $\widetilde{c}_t$
the tracked progress level.
We set $\widetilde{c}_t=c_t$ unless atomic-task
completion is confirmed, in which case it is reset
to zero. The task index advances only if another
atomic task remains.
The updated progress potential is
\begin{equation}
\Phi_{t+1}
=
\frac{z_{t+1}-1+\widetilde{c}_t/B}{K}.
\label{eq:pbsr_potential}
\end{equation}
The potential combines the atomic-task index with
within-task progress, with the latter scaled by $1/B$.

\paragraph{Shaping reward.}
For each valid non-terminal PCC update, we compute
\begin{equation}
r_t^{\mathrm{PBSR}}
=
\gamma\Phi_{t+1}-\Phi_t,
\label{eq:pbsr_definition}
\end{equation}
where $\gamma$ is the discount factor and $\Phi_t$
is the previously stored potential.
The stored potential is initialized to zero when
the task tracker is reset and replaced by
$\Phi_{t+1}$ after each valid update.
The shaping contribution is zero on terminal
transitions, including environment timeouts,
and whenever no valid update is available.
The training reward is
\begin{equation}
r_t
=
r_t^{\mathrm{env}}
+
\alpha_n r_t^{\mathrm{PBSR}},
\label{eq:pbsr_combined_reward}
\end{equation}
where $r_t^{\mathrm{env}}$ is the sparse task reward
and $\alpha_n$ controls the shaping contribution
at training iteration $n$.
The shaping reward can be positive or negative
and is not clipped.

\paragraph{Shaping schedule.}
We linearly anneal the shaping coefficient:
\begin{equation}
\alpha_n
=
\alpha_0
\left[
1-(1-\rho)
\operatorname{clip}\!\left(
\frac{n-n_s}{n_e-n_s},\,0,\,1
\right)
\right],
\label{eq:pbsr_annealing}
\end{equation}
where $\alpha_0$ is the initial coefficient,
$n_s<n_e$ specify the annealing interval,
$\rho \in [0,1]$ is the final retained fraction,
and $\operatorname{clip}(x,0,1)$ clamps $x$ to $[0,1]$.
Setting $\rho=0$ removes PBSR after annealing.

PBSR modifies the reward used to compute TD residuals,
whereas SAPO regulates their cross-boundary propagation
during advantage estimation.

\subsection{Additional Ablations}
\label{app:gaeseg}

\mypara{Sensitivity to the SAPO Gating Coefficient.}
We examine the effect of the SAPO gating coefficient $\kappa$ on LIBERO-LONG under the same configuration as the component ablations. 
\begin{wraptable}{r}{0.35\textwidth}
    \centering
    \small
    \setlength{\tabcolsep}{8pt}
    \renewcommand{\arraystretch}{1.12}

    \caption{\textbf{Effect of the SAPO gating coefficient.}}
    \label{tab:kappa_ablation}

    \begin{tabular}{cc}
        \toprule
        $\kappa$ & \textbf{Success Rate} \\
        \midrule
        0.0 & 87.2\% \\
        0.3 & \textbf{87.8\%} \\
        0.5 & 85.0\% \\
        0.8 & 81.2\% \\
        1.0 & 78.2\% \\
        \bottomrule
    \end{tabular}
\end{wraptable}
Here, $\kappa$ controls trace propagation across boundaries preceding failed subtasks, while gates preceding successful subtasks remain fixed at one. 
Setting $\kappa=1$ recovers standard GAE propagation, whereas $\kappa=0$ blocks cross-boundary trace propagation from failed subtasks without removing the one-step value bootstrap.
As shown in Table~\ref{tab:kappa_ablation}, the highest observed success rate is $87.8\%$ at $\kappa=0.3$, improving over the unattenuated baseline ($78.2\%$) by 9.6 percentage points. Performance decreases to $85.0\%$ and $81.2\%$ as $\kappa$ increases to 0.5 and 0.8, respectively. Hard trace blocking ($\kappa=0$) also performs well, achieving $87.2\%$, only 0.6 percentage points below $\kappa=0.3$. These results support the use of soft attenuation of cross-boundary trace propagation from failed subtasks in this setting.

\subsection{Training Details}
\label{app:sim-detail}
\mypara{Training Setup.} For all LIBERO and real-world experiments, we train PCC offline on annotated demonstrations. PCC uses a frozen SigLIP visual encoder with approximately 400M parameters, initialized from the $\pi_{0.5}$ visual backbone, together with lightweight trainable temporal and prediction modules.  
For data annotation, we adapt the annotation pipeline
of AtomicVLA~\citep{zhang2026atomicvlaunlockingpotentialatomic} to generate
fine-grained annotations for demonstration trajectories.
We annotate 150 trajectories for LIBERO and
50 trajectories per real-world task, which are
subsequently used for offline PCC training. The trained PCC is kept fixed during RL post-training.

All RL post-training experiments are implemented
in the RLinf framework.
All training is performed on eight NVIDIA H200 GPUs,
while inference is conducted on a single NVIDIA
GeForce RTX 5090 GPU.

\begin{figure}[t]
    \centering

    \begin{subfigure}[t]{0.32\linewidth}
        \centering
        \includegraphics[width=\linewidth]{
            figure/libero10_success_once_first_200_steps.pdf
        }
        \caption{LIBERO-LONG success rate.}
        \label{fig:libero10-success-once}
    \end{subfigure}
    \hfill
    \begin{subfigure}[t]{0.32\linewidth}
        \centering
        \includegraphics[width=\linewidth]{
            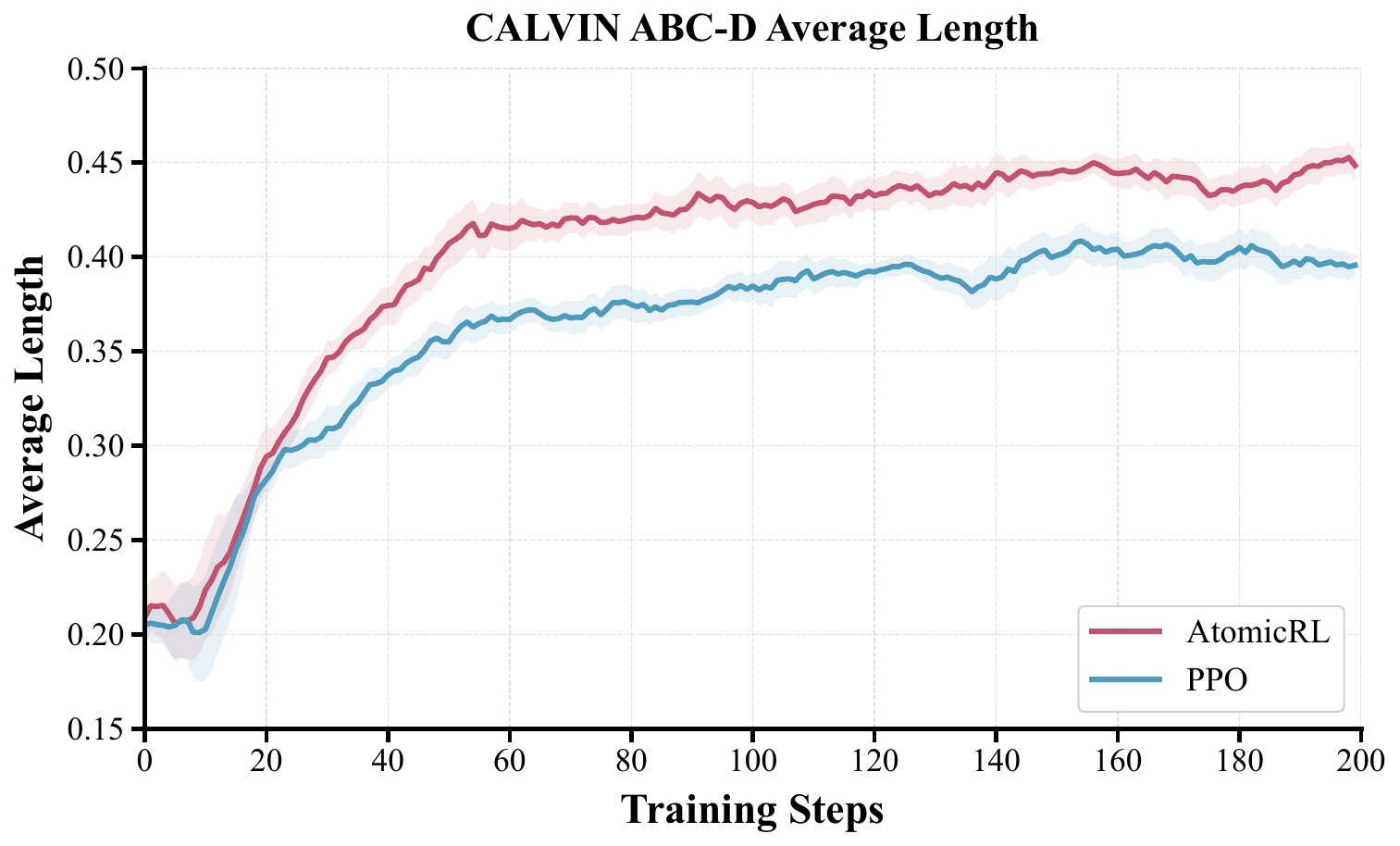
        }
        \caption{CALVIN average length.}
        \label{fig:calvin-success-once}
    \end{subfigure}
    \hfill
    \begin{subfigure}[t]{0.32\linewidth}
        \centering
        \includegraphics[width=\linewidth]{
            figure/calvin_success_subtask5_cond_first_200_steps.pdf
        }
        \caption{CALVIN task-5 success rate.}
        \label{fig:calvin-subtask5-cond}
    \end{subfigure}

    \caption{
        Training curves on LIBERO-LONG and CALVIN ABC-D.
    }
    \label{fig:training-curves-appedx-1}
\end{figure}

\begin{figure}[t]
    \centering

    \begin{subfigure}[t]{0.42\linewidth}
        \centering
        \includegraphics[width=\linewidth]{
        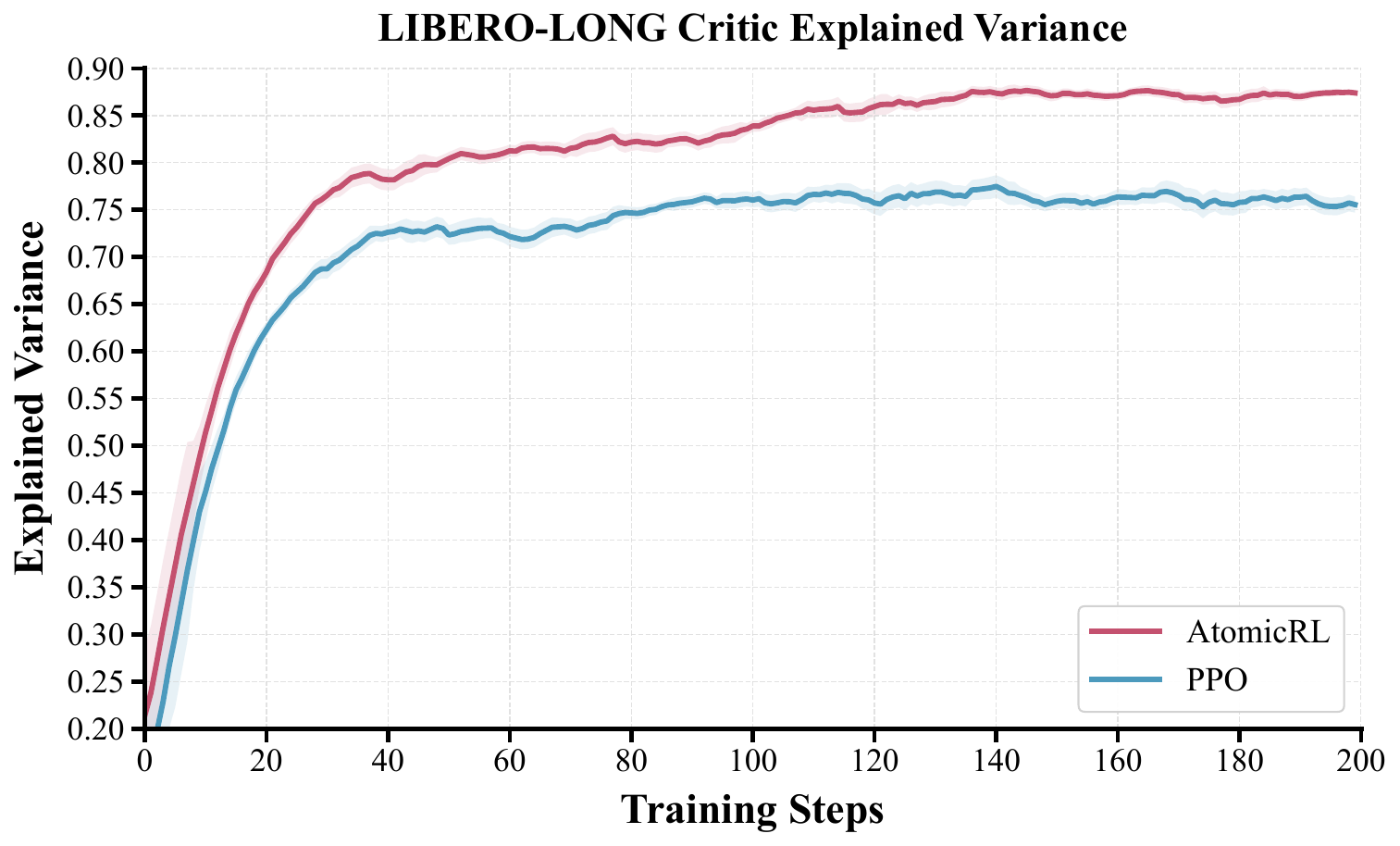
        }
        \caption{LIBERO-LONG critic explained variance.}
        \label{fig:task1-success-once}
    \end{subfigure}
    \hfill
    \begin{subfigure}[t]{0.42\linewidth}
        \centering
        \includegraphics[width=\linewidth]{
            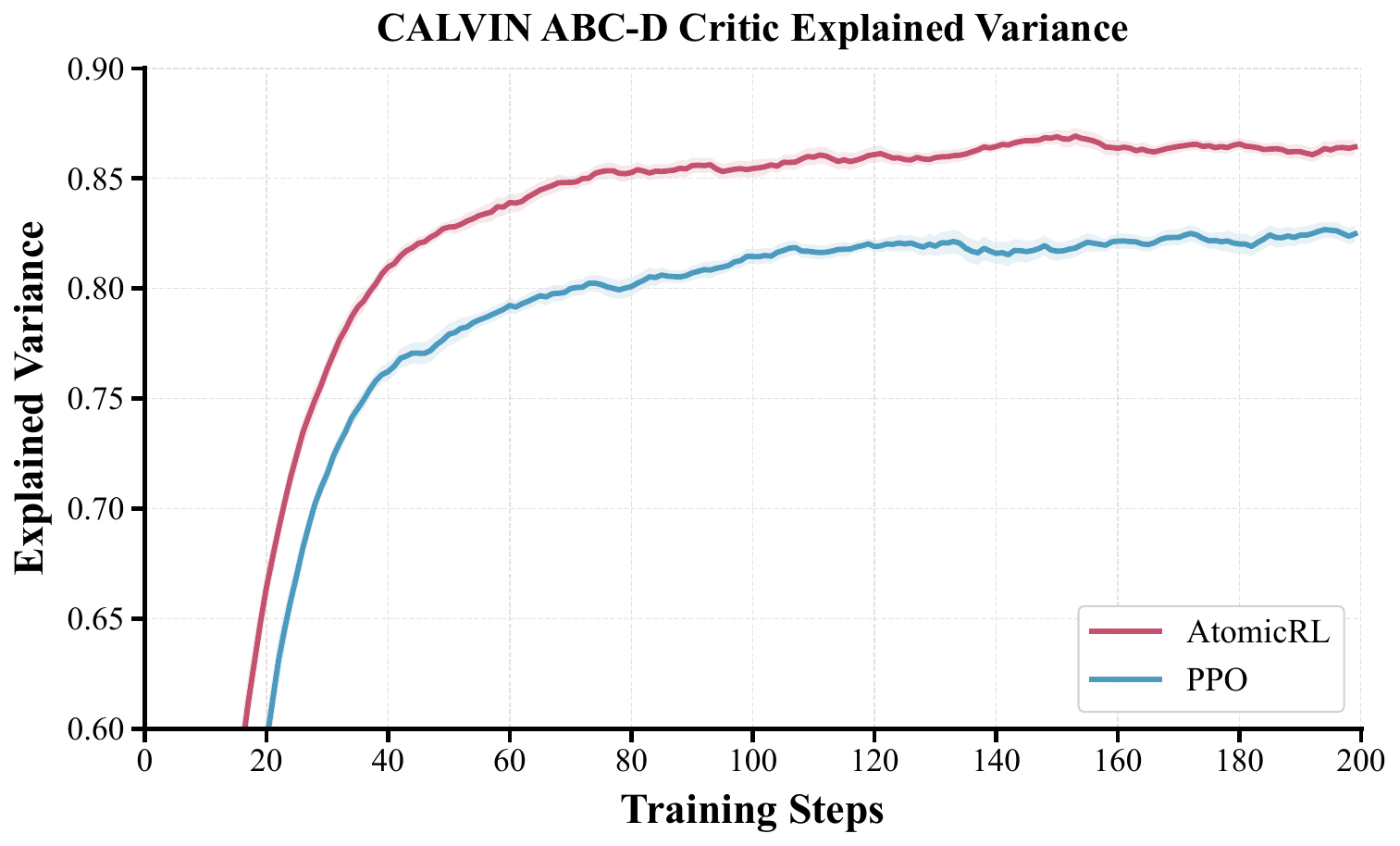
        }
        \caption{CALVIN critic explained variance.}
        \label{fig:task2-success-once}
    \end{subfigure}
    \caption{
        Critic explained variance.
    }
    \label{fig:critic_explained_variance}
\end{figure}

\begin{figure}[t]
    \centering

    \begin{subfigure}[t]{0.32\linewidth}
        \centering
        \includegraphics[width=\linewidth]{
            figure/bread_placement_success_rate.pdf
        }
        \caption{BP success rate.}
        \label{fig:task1-success-once}
    \end{subfigure}
    \hfill
    \begin{subfigure}[t]{0.32\linewidth}
        \centering
        \includegraphics[width=\linewidth]{
            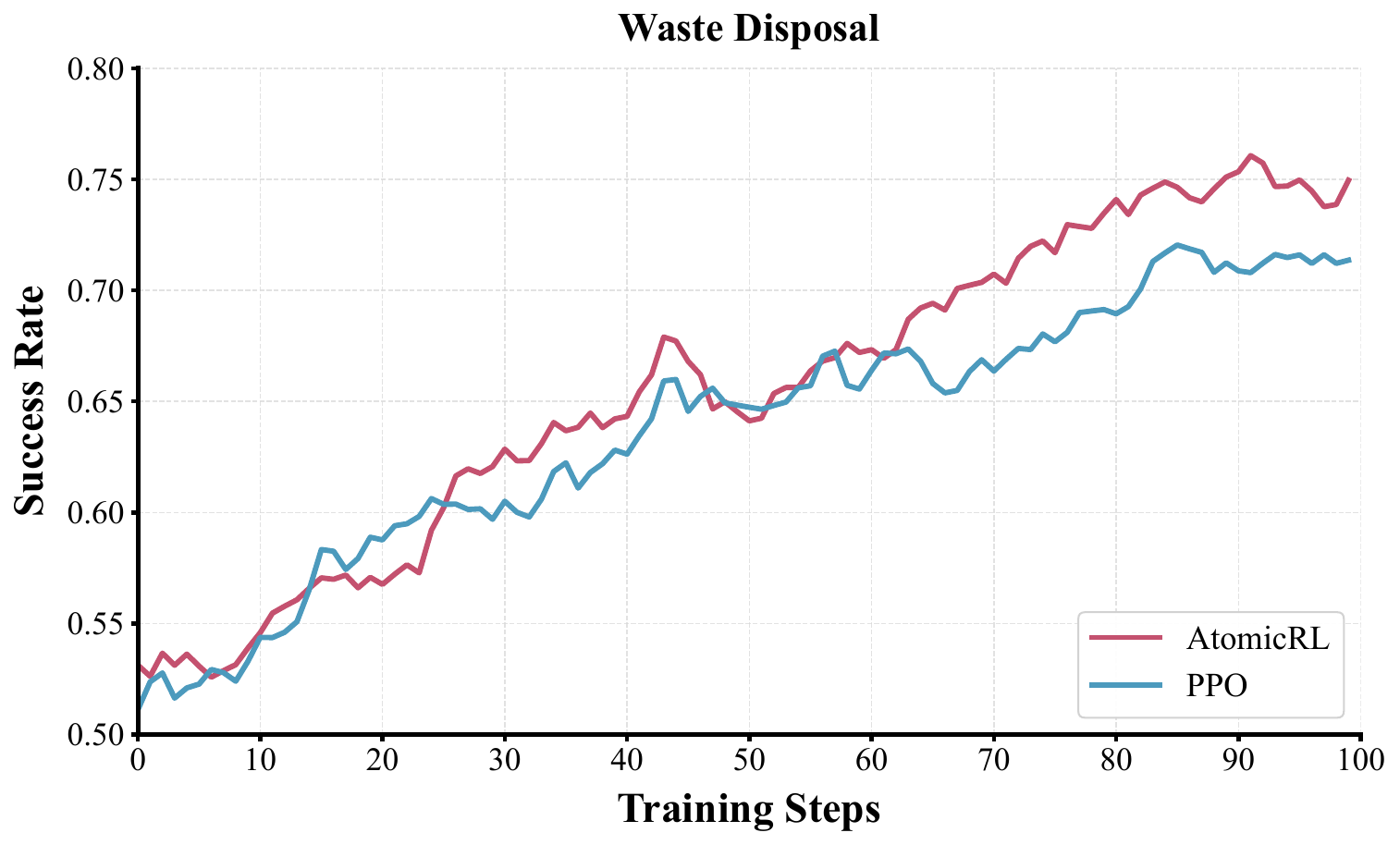
        }
        \caption{WD success rate.}
        \label{fig:task2-success-once}
    \end{subfigure}
    \hfill
    \begin{subfigure}[t]{0.32\linewidth}
        \centering
        \includegraphics[width=\linewidth]{
            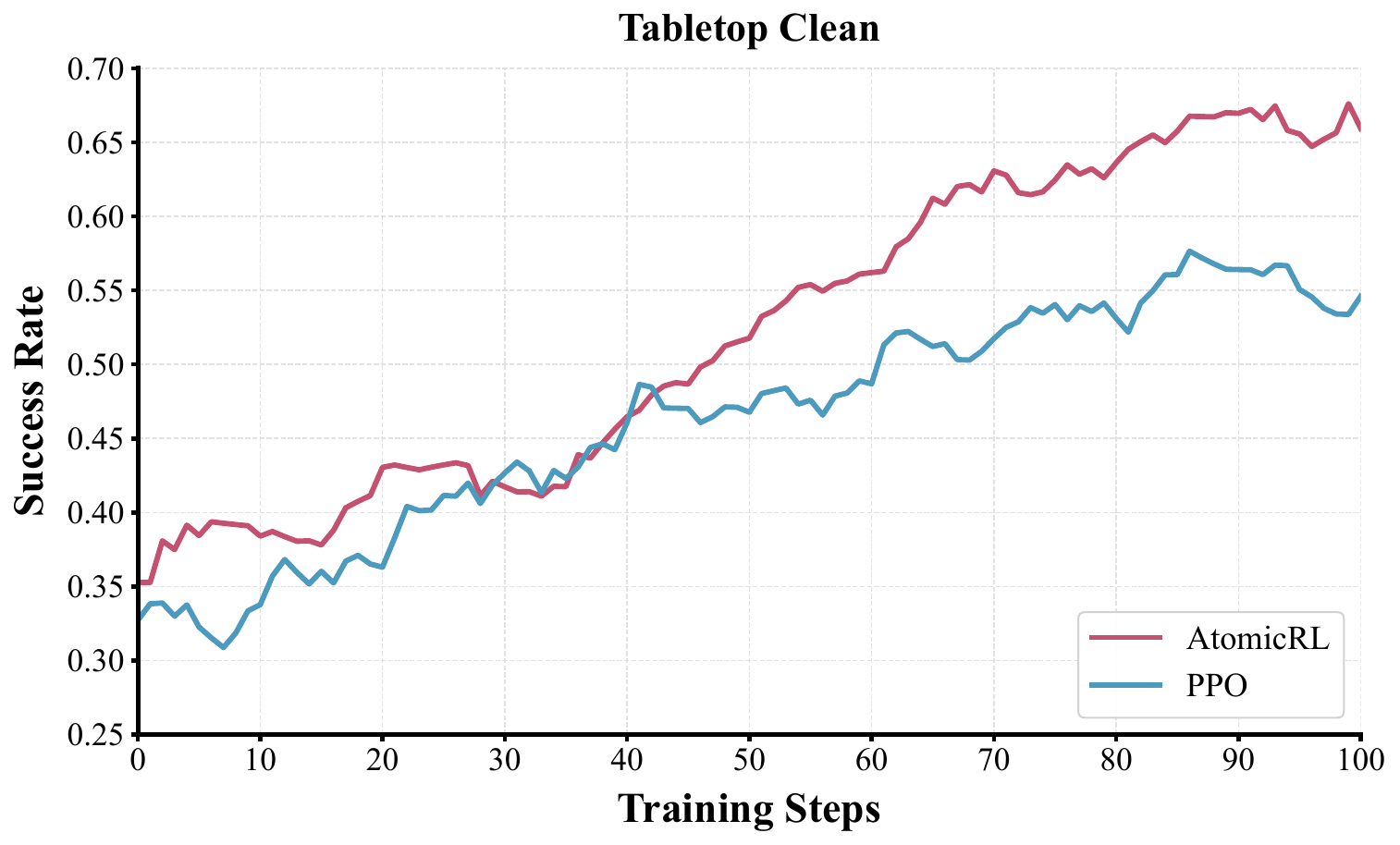
        }
        \caption{TC success rate.}
        \label{fig:task3-success-once}
    \end{subfigure}

    \caption{
        Training curves on Real-world robot.
    }
    \label{fig:training-curves_real-2}
\end{figure}

\subsection{Additional Results}
\mypara{Detail Results on benchmarks.} Fig.~\ref{fig:training-curves-appedx-1} and Fig.~\ref{fig:critic_explained_variance} show the training dynamics and value-function fitting of AtomicRL. On both LIBERO-LONG and CALVIN ABC-D, AtomicRL improves more rapidly during early training and consistently outperforms PPO throughout subsequent training. Meanwhile, the critic explained variance in the later stages of training is higher than PPO by approximately 12\% on LIBERO and 5\% on CALVIN, indicating that the value function explains a larger fraction of the variance in the training targets. These trends are consistent with the design of SAPO and BATE: SAPO attenuates the cross-boundary propagation of TD residuals from failed atomic subtasks to completed prefixes, improving temporal credit assignment and reducing interference from downstream failures with previously successful behaviors; BATE concentrates sampling and exploration resources on insufficiently mastered bottleneck subtasks. The concurrent improvement in task performance and value-function fitting further supports the effectiveness of atomic-task-structured optimization for long-horizon policy learning.

\mypara{Detail Results on real-world robot.} 
As shown in Fig.~\ref{fig:training-curves_real-2}, when online RL is performed in the Real2Sim-reconstructed environments, AtomicRL exhibits better training performance than PPO across all three real-world tasks. The advantage is particularly pronounced during early training on Bread Placement (BP), which consists of six atomic subtasks: PPO remains around 50\% success over the first 40 training steps, whereas AtomicRL reaches 60\% at around step 30 and continues to improve thereafter. This gap reflects the difficulty of learning long task chains under sparse feedback and is consistent with the design of AtomicRL, which attenuates cross-boundary credit interference from downstream failures and thereby makes better use of successful prefixes in failed trajectories. AtomicRL also achieves higher late-stage training success rates on Waste Disposal (WD) and Tabletop Clean (TC), indicating that its training advantage is not limited to a single task. Combined with the simulation--real-world performance correlation analysis in Fig.~\ref{fig:real_sim_transfer}, these training curves provide an indirect reference for the trend of real-world policy improvement.

\subsection{Details of automated Real2Sim2Real pipeline}
\label{app:real2sim2real}
\begin{figure}[t]
    \centering
    \includegraphics[
        width=\linewidth
    ]{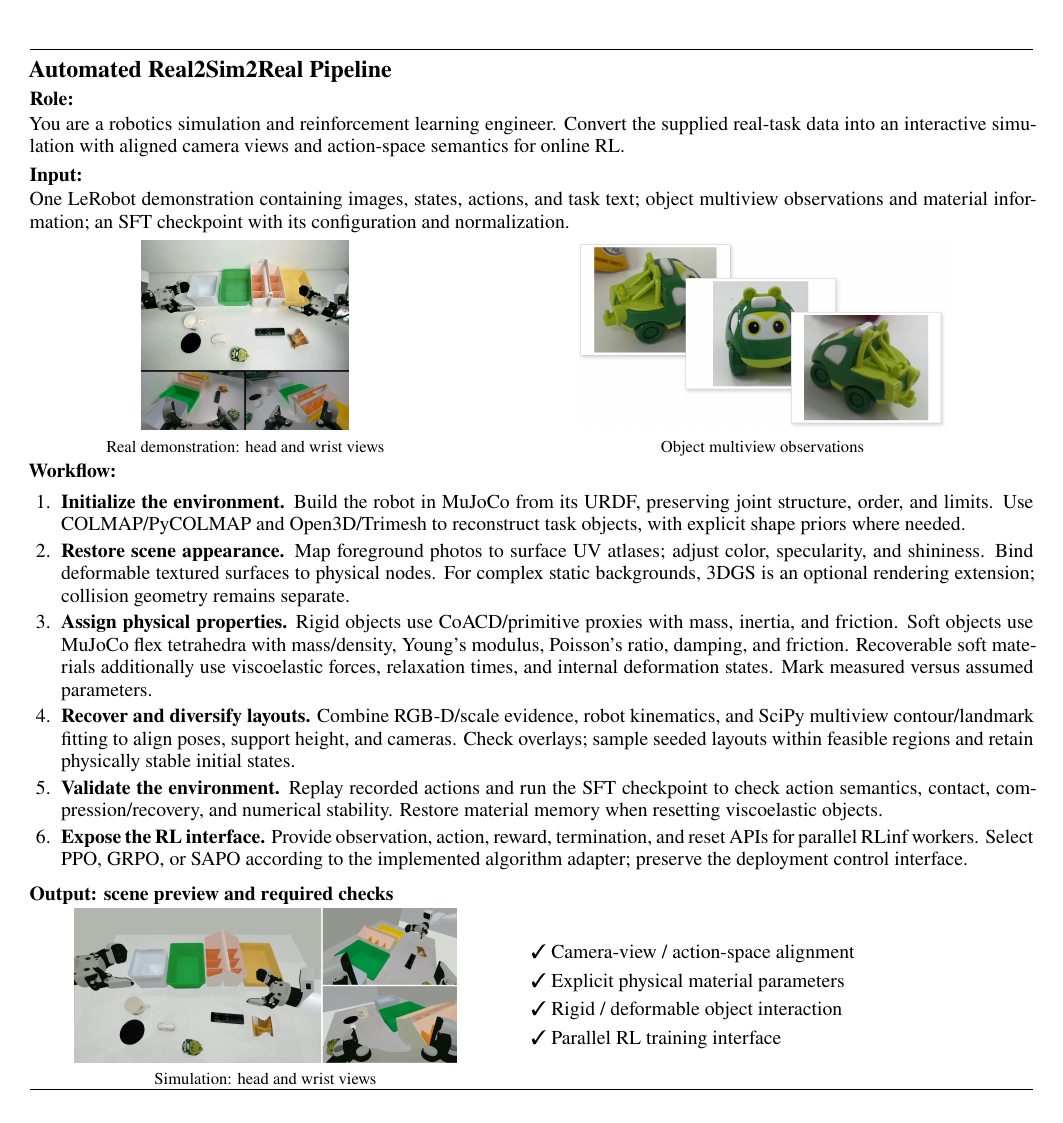}
    \caption{Automated Real2Sim2Real workflow with essential
    tool calls and a real-to-simulation scene example.}
    \label{fig:real2sim2real_pipeline}
\end{figure}

As shown in Fig.~\ref{fig:real2sim2real_pipeline},
we develop an automated Real2Sim2Real workflow,
with GPT-6 Astra assisting tool orchestration
and code implementation.
The workflow connects real-task reconstruction,
simulation-based online RL in RLinf~\citep{yu2025rlinfflexibleefficientlargescale}, and real-robot
deployment, with the key steps and validation
requirements summarized in the figure.

\subsection{Additional Visualizations}
\begin{figure}[h]
    \centering
    \vspace{-5pt}
    \includegraphics[width=1.0\textwidth]{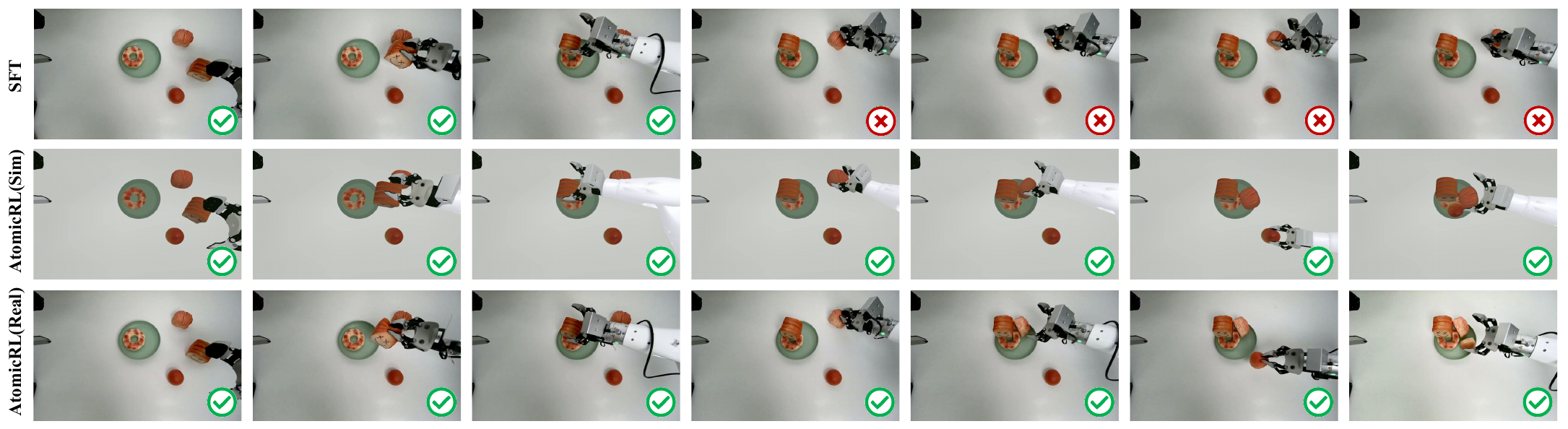}
    \caption{Demonstrations show the execution process of SFT and AtomicRL.}
    \label{bread-vis} 
    \vspace{-12pt} 
\end{figure}

We provide additional visualizations of policy execution in simulation and the real world. As shown in Fig.~\ref{bread-vis}, SFT completes the initial actions but fails at a later stage, whereas AtomicRL successfully completes the illustrated sequence in both simulation and real-world deployment. 

